\documentclass[conference]{IEEEtran}

\usepackage{amsmath}
\usepackage{times}
\usepackage{epsfig}
\usepackage{graphicx}
\usepackage{amsmath}
\usepackage{amssymb}
\usepackage{multirow}
\usepackage{float}
\usepackage{subfigure}

\usepackage{xspace}
\newcommand{\eg}{\emph{e.g.}\@\xspace}
\newcommand{\ie}{\emph{i.e.}\@\xspace}
\newcommand{\etal}{\emph{et al.}\@\xspace}
\DeclareMathOperator*{\maxpool}{MP}

\newcommand{\Cout}{C_{\mathrm{out}}}
\newcommand{\Cin}{C_{\mathrm{in}}}

\usepackage{todonotes}
\usepackage{regexpatch}
\makeatletter
\xpatchcmd{\@todo}{\setkeys{todonotes}{#1}}{\setkeys{todonotes}{inline,#1}}{}{}
\makeatother

\title{Object Detection in 3D Point Clouds via Local Correlation-Aware Point Embedding}

\author{
\IEEEauthorblockN{\\Chengzhi Wu}
\vspace{0.2cm}
\IEEEauthorblockA{Institute for Anthropomatics\\ and Robotics,\\
Karlsruhe Institute of Technology,\\ Karlsruhe, Germany  \vspace{-0.54cm}}
\and
\IEEEauthorblockN{Julius Pfrommer \\ Jürgen Beyerer}
\vspace{0.4cm}
\IEEEauthorblockA{Fraunhofer Center for Machine Learning,\\
Fraunhofer IOSB,\\Karlsruhe, Germany 
\vspace{-0.54cm}}
\and
\IEEEauthorblockN{Kangning Li \\ Boris Neubert}
\vspace{0.2cm}
\IEEEauthorblockA{Institute for Visualization\\ and Data Analysis,\\
Karlsruhe Institute of Technology,\\ Karlsruhe, Germany \vspace{-0.54cm}}
}

\begin{document}

\maketitle
\begin{abstract}
We present an improved approach for 3D object detection in point clouds data based on the Frustum PointNet (F-PointNet). Compared to the original F-PointNet, our newly proposed method considers the point neighborhood when computing point features. The newly introduced local neighborhood embedding operation mimics the convolutional operations in 2D neural networks. Thus features of each point are not only computed with the features of its own or of the whole point cloud, but also computed especially with respect to the features of its neighbors. Experiments show that our proposed method achieves better performance than the F-Pointnet baseline on 3D object detection tasks.
\end{abstract}


\section{Introduction}
\label{sec:intro}
In computer vision, the task of 3D object detection in point cloud data is of central importance to various applications including robotics, autonomous driving, and virtual/augemented reality. 
Compared to remarkable progress made on 2D detection and segmentation with various neural networks \cite{Girshick2015FastRCNN,Ren2015FasterRCNN,He2017MaskRCNN,Liu2016SSD}, 3D detection is relatively less explored. 
In contrast to 2D images that have a dominant representation of pixel arrays, which are always well aligned and perfect for applying convolution operations, 3D point clouds are usually irregular, unordered and potentially sparse. Most existing deep learning-based 3D object detection methods convert point clouds into regular forms including images and voxels. 
But these representations obscure the natural invariance of 3D shapes under geometric transformations. They also lead to difficult trade-offs between sampling resolution and network efficiency \cite{Fan2017PointSetGene}. 

A number of papers have been proposed to apply deep learning techniques directly on raw point clouds without conversions. PointNet \cite{Qi2017PointNetDL} uses max-pooling as a symmetric function to deal with the unordered nature of point cloud data. Each point is represented by its 3D coordinates with additional features computed by subsequent networks. However, the originally proposed PointNet only took global information into consideration, \ie the pooling operation was applied on the whole set of the input point cloud. In the subsequent work PointNet++ \cite{Qi2017PointNetDH} local information has been considered by applying multi-scale grouping or multi-resolution grouping. Still, these region-wise features cannot represent points by means of their local correlations well. 

Following this work, Frustum-PointNet \cite{Qi2018frustum} was proposed for amodal 3D object detection with pre-segmented object frustums. Other methods, such as RoarNet \cite{Shin2018RoarNetAR} and PointRCNN \cite{Shi2018PointRCNN3O} have been proposed by using PointNet as their backbone networks to extract point-wise features for different learning tasks. Applying the idea of spatial sampling, there are also methods that divide the whole point clouds into voxel blocks \cite{Zhou2018VoxelNetEL, Yan2018SECONDSE} or pillars \cite{Lang2019PointPillarsFE}, embedded features of the input objects are extracted by applying PointNet on each of the voxel blocks or pillars. However, those methods may encounter the problem of splitting objects due to inappropriate discretizations when partitioning. This is especially problematic for unevenly sampled environments.

There are several problems when applying deep learning methods directly on raw point cloud data: (i) operations on a point cloud should be independent of the input order of the points; (ii) the representation of a point cloud should be invariant to certain transformations, including simple affine and geometric transformations; and (iii) local correlation around each point should be considered to capture shape information, \ie a local correlation-based information gathering operation which mimics the convolution operations in 2D CNNs should be designed.
Most forementioned state-of-the-art methods only aim at the first problem.
To better tackle the second and third problems, based on the work of PointNet and Frustum-PointNet, we propose a framework for 3D object detection that especially addresses the task of gathering local neighborhood information in point clouds. The point neighborhoods are embedded in their final representations to achieve an improved segmentation and detection performance.


\begin{figure*}[ht]
\begin{center}
    \includegraphics[width=1.0\linewidth]{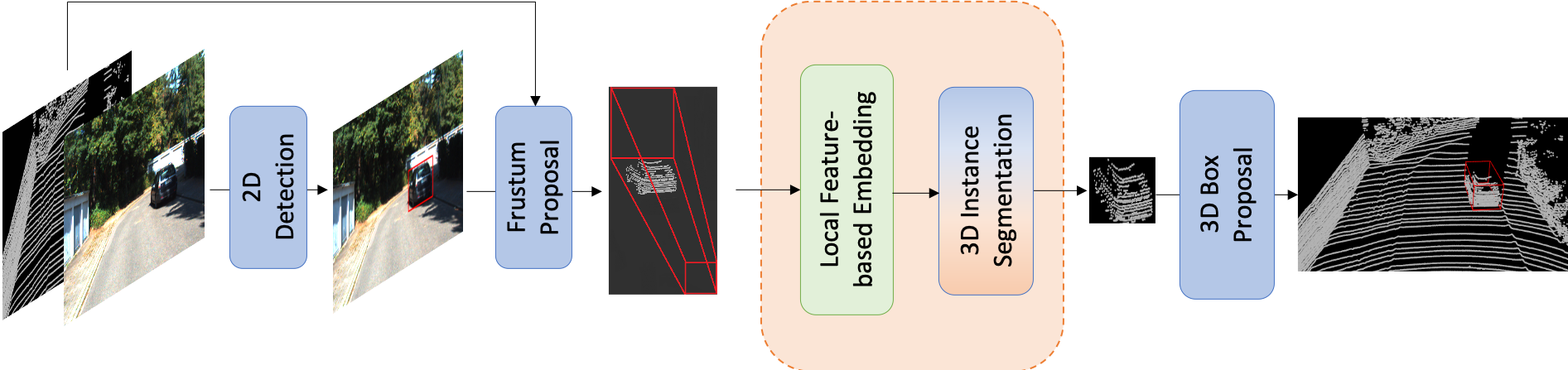}
\end{center}
    \caption{The framework of our method, some parts of it are identical to the F-PointNet (colorized in blue) except for the segmentation part (colorized in orange). We add an additional local feature embedding block forehead (colorized in green), while the fully connected operations in the original 3D instance segmentation network may also be replaced by the proposed local feature-based embedding operations.}
\label{fig:framework}
\end{figure*}

The key contribution of this work is summarized as follows:
\begin{itemize}
\setlength{\topsep}{0pt}
\setlength{\itemsep}{2pt}
\setlength{\parsep}{0pt}
\setlength{\partopsep}{0pt}
\setlength{\parskip}{0pt}
    \item We introduce a novel local information gathering operation to embed local correlations for all the points in a point cloud.
    \item A new embedding block is constructed for prior feature extraction and embedding, while the newly introduced operation can also replace the fully connected layers in the original PointNet.
    \item With the newly proposed framework, we achieve superior 3D detection performance on all categories on the KITTI dataset, compared to F-PointNet as our baseline.
\end{itemize}

\newpage
This paper is structured as follows. In Section 2, we briefly review related work in the scope of 3D object detection via deep learning methods. The local information gathering operation and the new embedding block are introduced in Section 3. Section 4 gives experimental results along with relevant ablation studies. A conclusion and future outlook are presented in Section 5.

\vspace{5pt}
\section{Related Work}
In this section, we briefly review existing deep learning-based 3D object detection methods for point cloud data.
Those methods either convert point clouds into images/voxels for learning, or do direct learning on the points.

\textbf{Image/Voxel-based methods:} In order to apply classical 2D/3D convolutional neural networks, some methods convert the point clouds into more regular representations by projecting them into images or discretizing them into voxel grids. Certain perspective RGB-D images was used for 3D amodal detection in \cite{Deng2017AmodalDO}. MV3D \cite{Chen2017Multiview3O} projects point clouds to bird's eye view (BEV) and uses Faster-RCNN \cite{Ren2015FasterRCNN} to learn features from point clouds for object detection. AVOD \cite{Ku2018Joint3P} proposes a feature fusion Region Proposal Network (RPN) that utilizes multiple modalities to produce high recall region proposals from BEV. PIXOR \cite{Yang2018PIXORR3} extends their work by exploiting the BEV representation in a more efficient way with a proposal-free object detector. Front-view images have also been used in some methods \cite{Liang2018DeepCF, Li2016VehicleDF}. 3D-SSD \cite{Luo20203DSSDLH} uses  RGB images and depth images to generate 3D bounding boxes directly. On the voxel format side, Qi \etal explores the representations of volumetric objects \cite{Qi2016VolumetricAM} while Li utilizes it for vehicle detection in point clouds \cite{Li20173DFC}. Voxnet \cite{Maturana2015VoxNetA3} and 3DBN \cite{Li20193DBN} also directly discretize the point cloud scenes into voxels to apply similar operations. DSS \cite{Song2016DeepSS} applies 3D CNNs to equally divided volumetric space of grids in its 3D amodal Region Proposal Network as well. 
However, when projecting or quantifying the point cloud data, all of the above methods suffer severely from the problem of information loss including 3D-to-2D transition information loss and data space discretization information loss, whose influence are always inneglectable.

\textbf{Point-based methods:} Due to the above-mentioned disadvantages, researchers have started to work on directly processing raw point clouds data. PointNet  \cite{Qi2017PointNetDL} is the first proposed deep learning-based method to directly take 3D raw point clouds as input. It uses max pooling as a symmetric function to cope with the unordered nature of point cloud data. Subsequent work PointNet++ \cite{Qi2017PointNetDH} adds one more step of grouping and extracting local features. Using it as a backbone network for 3D object detection tasks Frustum-PointNet \cite{Qi2018frustum} was proposed, which we use as our baseline.
VoxelNet \cite{Zhou2018VoxelNetEL} and Second \cite{Yan2018SECONDSE} divide space into voxels to compute local features for the point clouds. Note that in contrast to the converting-to-voxels methods mentioned above voxels are used as a grouping strategy to group and embed new features for the points, which may be regarded as a variant of PointNet++ \cite{Qi2017PointNetDH}. Similar methods include PointPillars \cite{Lang2019PointPillarsFE} that divides space into pillars, and RSNet \cite{Huang2018RecurrentSN} that uses a sliding block to extract features.
PU-Net \cite{Yu2018PUNetPC} uses an upsampling strategy to learn multi-level features of point clouds via a multi-branch convolution unit. SplatNet \cite{Su2018SPLATNetSL} stacks bilateral convolution layers to construct its network. Inspired from Faster-RCNN \cite{Ren2015FasterRCNN}, another end-to-end framework PointRCNN \cite{Shi2018PointRCNN3O} has also been proposed to generate 3D proposals from raw point clouds in a bottom-up manner.
Following the idea of PointNet++ \cite{Qi2017PointNetDH}, DGCNN  \cite{Wang2019DynamicGC} firstly tried to design local convoluion operations to convolve local information. The proposed EdgeConv operation has also been adopted in other researches like unsupervised multi-task learning \cite{Hassani2019UnsupervisedMF}. Similar ideas have been proposed in other works, \eg KCNet \cite{Shen2018MiningPC} uses kernel correlation (KC) operations, and PointCNN \cite{Li2018PointCNN} uses $\mathcal{X}$-Transformation operations for local-based point cloud feature learning.

\vspace{5pt}
\section{Methodology}


Our approach extends the main structure of F-PointNet with a novel embedding block and segmentation network. Figure \ref{fig:framework} shows the overall network architecture.

\subsection{Point-Neighborhood Embedding Operation}
\label{sec:knn-embedding}

In this section, a KNN-based embedding operation for local point neighborhoods is introduced. It generalizes the idea of EdgeConv \cite{Wang2019DynamicGC} to capture local geometric structures of point clouds. 
A point cloud is parameterized as $P=\{p_1, p_2, \dots, p_N \}$.
Every point $p_i$ is a vector of point attributes, such as the location and additional attributes, such as color, intensity, classification label, and so on. We say attributes for the original data associated with each point and features for higher-order representations and embeddings computed from the attributes. For some of the point attributes, the difference between points is meaningful. For example, the difference between the positions in Cartesian coordinates gives the distance vector. For other attributes, \eg a nominal classification label, there is no well-founded difference defined.
To differentiate between the two kinds of point attributes, each point $p_i$ decomposes into two vectors $p_i=(c_{i},v_{i})$. The vector $c_{i}$ contains the attributes for which differences between points can be computed. The vector $v_{i}$ contains the attributes that are not comparable.

\begin{figure}[t]
\begin{center}
    \includegraphics[width=0.9\linewidth]{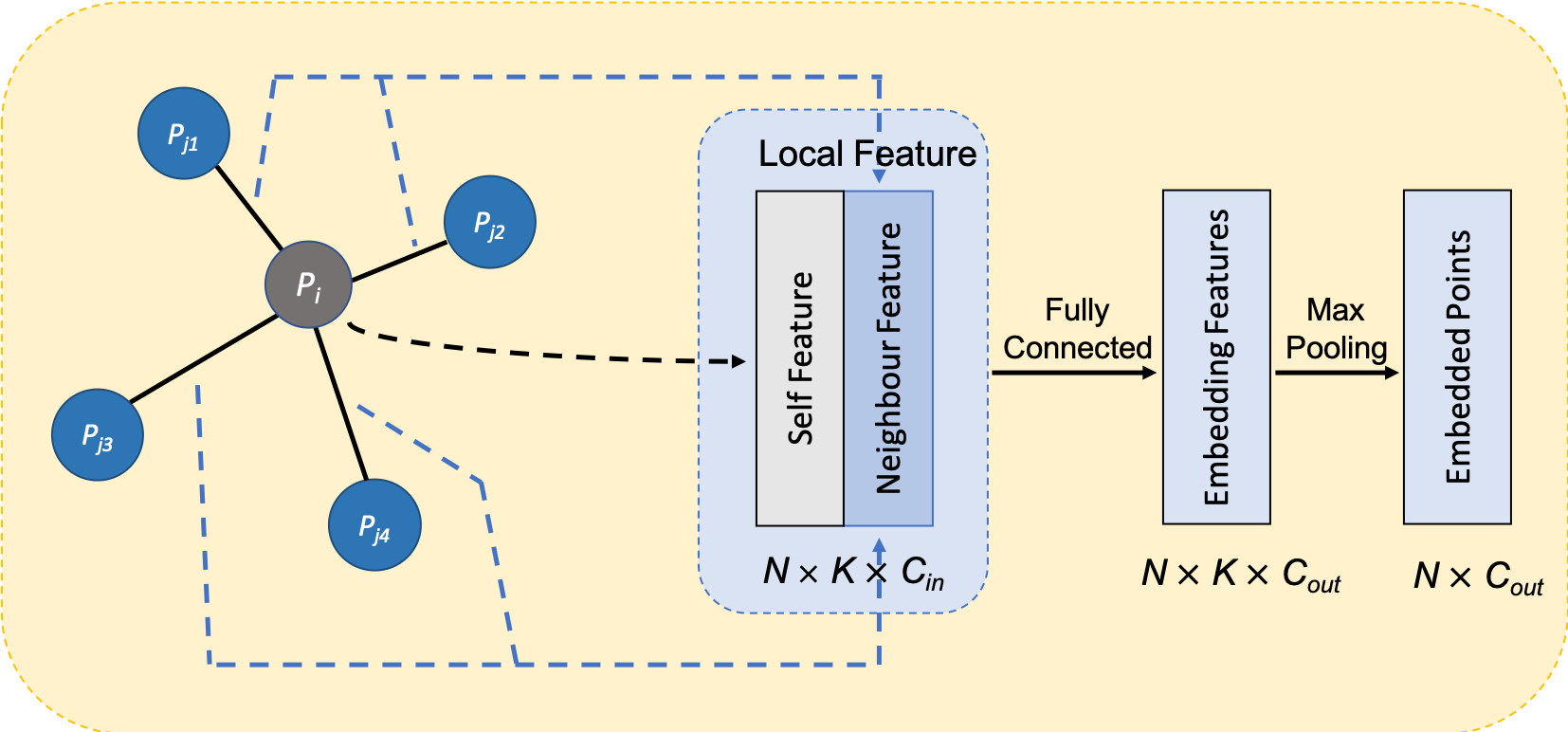}
\end{center}
    \caption{Illustration of one local feature-based embedding operation. $K$ nearest neighbours of each point are selected to contribute in computing its local features. Processing the input features through one fully connected layer followed by one pooling layer, finally we get the embedded output features of points, in which the local correlation information are implicitly represented. $N$ stands for the number of points, $K$ stands for the number of neighbour points selected, $\Cin, \Cout$ stand for the channel numbers of input features and embedded output features respectively.
    }
\label{fig:oneLayer}
\end{figure}

For each point $p_i$, we gather its $K$ nearest neighbours as a set $N_i=\{p_{j_1}, p_{j_2}, \dots, p_{j_K}\}$. 
The embedding of a local patch $(p_i, N_i)$ is computed as follows:
Every neighborhood relation between points $p_i$ and $p_j\in N_i$ is preprocessed through an input processing operator $\mathbb{D}$ into a data vector $\mathbb D(p_i, p_j)$ with dimension $\Cin$. 
The transformation into the embedding space for each neighbour is a function $f: \mathbb R^{\Cin} \to \mathbb R^{\Cout}$. In this paper, we train $f$ as a fully connected layer. The embedding for all neighbours in $N_i$ should not depend on their ordering. Therefore a final max-pooling operation is applied that also reduces the dimensionality of the overall embedding of the local patch information $\mathbb F: \mathbb R^{K\times \Cin} \to \mathbb R^{\Cout}$.
\begin{align}
    \mathbb{F}(p_i,N_i) &= \maxpool_{p_j\in N_i} f(\mathbb D(p_i, p_j))
    \label{equ:both}
\end{align}
Here, $\maxpool$ is the max-pooling operator which applies element-wise maximization between the embeddings for the individual neighbours $p_{j}$.
A detailed explanation of the neighborhood feature embedding operation is illustrated in Figure \ref{fig:oneLayer}.

\subsection{Point-Neighborhood Embedding Block}
\label{subsec:embedding block}

Multiple neighborhood embedding operations can be chained together to a sequence with an overall embedding. The index $o$ is used to distinguish between the individual operations $\mathbb{F}^o$. Of course, the input and output dimensionality between operations have to match with $\Cout^o = \Cin^{o+1}$. We denote the attributes of the original points as $p_i$ and the intermediary embedding vectors between operations by the index of the previous operation $p_i^o$.
Parameters are not shared between the embedding operations $\mathbb{F}^o$ at different steps.

Note that after each convolution and pooling layer, the representations of the points have changed from 3D coordinates into embedded vectors. Going deeper through the embedding network, the selection of KNN neighbors in every layer is based on the "new distance" from the previous embedding.
With this special property of our proposed sequence of embedding operations, points with similar features may be grouped together for the subsequent net layers.

\begin{figure}[t]
\begin{center}
    \includegraphics[width=1.0\linewidth]{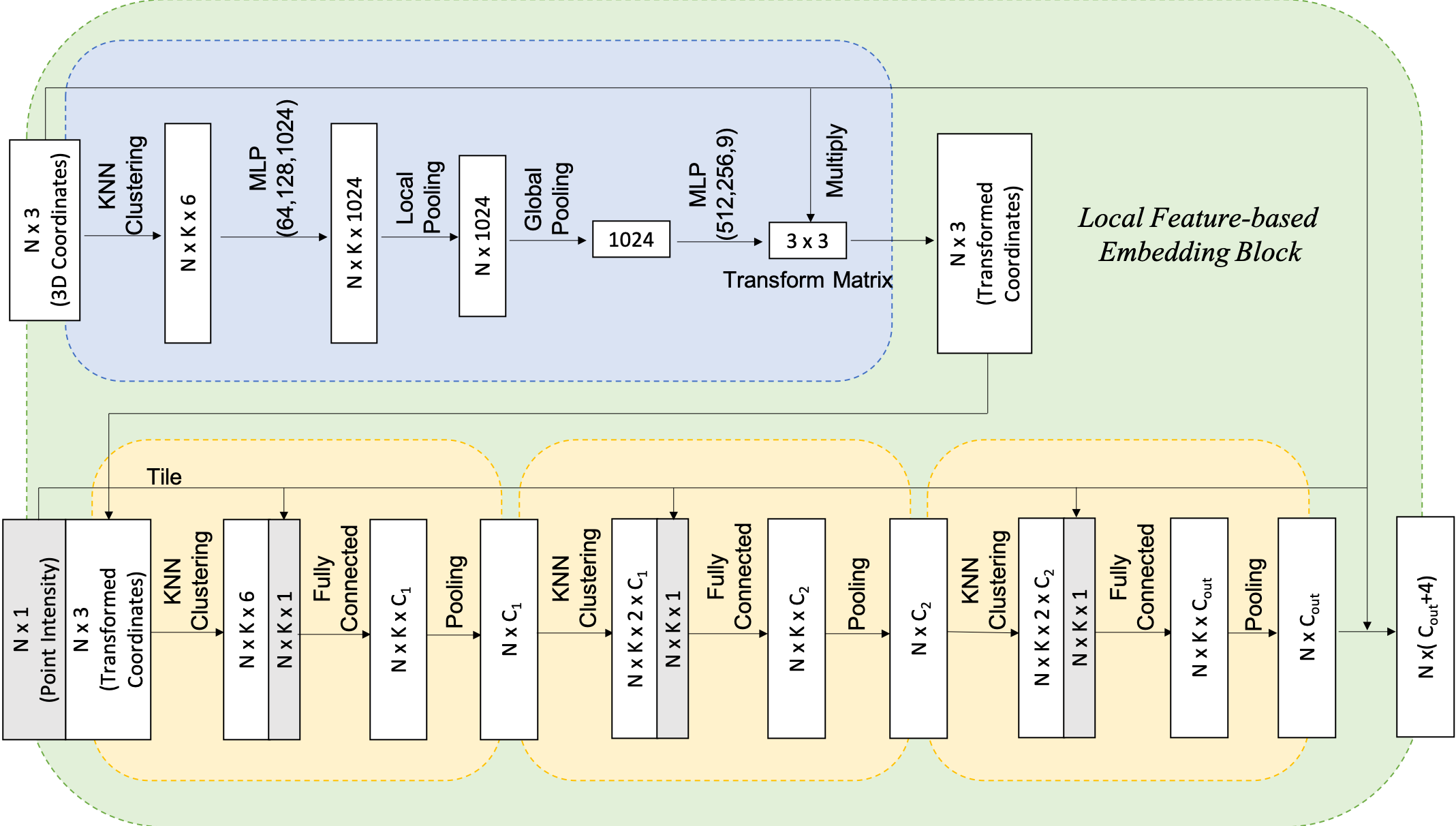}
\end{center}
    \caption{The network architecture of the point neighborhood embedding block. It consists of a spatial invariance transformation (ST) net (blue) and a sequence of neighborhood embedding operations (yellow). 
    }
\label{fig:embeddingBlock}
\end{figure}

Before the embedding operation sequence, a spatial invariance transform (ST) is added to normalize the frustrum orientations. PointNet \cite{Qi2017PointNetDL} initially introduced a spatial invariance transform prior to segmentation. In the framework of Frustrum-PointNet \cite{Qi2018frustum}, it was moved to a later stage for the computation of the bounding box center residual. But it was not used to improve the segmentation itself. Our experiments suggest that adding ST operation before the segmentation has a positive effect on segmentation results since it normalizes the frustum orientations. We thus apply the ST operation of \cite{Qi2017PointNetDL} prior to our KNN-based neighborhood embedding operations. At the end of the embedding block, the output from the last embedding operation is concatenated with the original input channels. The whole embedding block flow is illustrated in Figure \ref{fig:embeddingBlock}.

\renewcommand{\arraystretch}{1.3}
\begin{table*}[t]
\centering
\scalebox{1.3}{
\begin{tabular}{c||c|c|c|c}
\hline
Method & input processing operator $\mathbb{D}$ & ST sub-block & LFE operations & FCR operations \\  \hline
F-PointNet \cite{Qi2018frustum} & $\mathbb{D}(p_i, p_j)=p_i$ & at later stage & none & none \\  
F-PointNet++ \cite{Qi2018frustum} & $\mathbb{D}(p_i, p_j)=p_j$ & at later stage & none & none \\  
DGCNN \cite{Wang2019DynamicGC} & $\mathbb{D}(p_i, p_j)=(c_i, c_i-c_j)^{\top}$ & none & none & yes \\  \hline
Ours (EB) & $\mathbb{D}(p_i, p_j)=(c_i, c_i-c_j, \mathbb{O}(p_i))^{\top}$ & before embedding & yes & none \\  
Ours (EB+FCR) & $\mathbb{D}(p_i, p_j)=(c_i, c_i-c_j, \mathbb{O}(p_i))^{\top}$ & before embedding & yes & yes \\  \hline
\end{tabular}
}
\caption{Comparison between our method and other baseline methods regarding input processing operator $\mathbb{D}$ and other sub-blocks. EB stands for Embedding Block, ST stands for Spatial Transformation, LFE stands for Local Feature Embedding, FCR stands for Fully Connected layers Replaced, same below.}
\label{table:method comparison}
\end{table*}

In this paper, the input point clouds have a size of $N\times (3+1)$. Each of the $N$ points is represented by three Cartesian coordinate channels and one intensity channel (see Figure~\ref{fig:embeddingBlock}). Only the coordinate channels are used to apply the spatial transformation. The three subsequent embedding operations each produce an embedding of dimensionality $\Cout=64$. The data vector for each embedding operations is computed as 
\begin{equation}
    \mathbb{D}^o(p_i, p_j) =
    \begin{cases}
    (c_{i}, c_{i}-c_{j}, \mathbb O(p_i)), & o=0\\
    (p^{o-1}_{i}, p^{o-1}_{i}-p^{o-1}_{j}, \mathbb O(p_i)), & o\geq 1\,.
    \end{cases}
\end{equation}
The comparable features $c_i$ are the Cartesian point coordinates. Further, $\mathbb{O}(p_i)$ denotes attributes taken from the original point $p_i$. Non-comparable attributes $v_{i}$ may be included here. In our case, $v_i$ has the point intensity as its only member. It is possible to use neural network layers for the $\mathbb D^o$ as well and apply end-to-end training instead of feature engineering. But then prior knowledge about spatial relationships cannot be considered and the number of networks parameters is increased. Finally, the output feature map is concatenated with its corresponding 3D coordinate channels and the extra intensity channel to compose a $N(4+\Cout)$ embedded feature map, in which local correlations in the frustum point clouds have been implied.

Note that our method includes the original PointNet model \cite{Qi2017PointNetDL} as a special case with $\mathbb D(p_i, p_j) = p_i$. Another common choice for $\mathbb{D}$ comes from DGCNN \cite{Wang2019DynamicGC} with $\mathbb{D}(p_i, p_j)=(c_{i}, c_{i}-c_{j})^{\top}$, in which only comparable features are considered. Although sharing a similar idea from EdgeConv, our approach is different from it in: 1) the EdgeConv operator does not include the point itself into the set of neighbours, while we do, which ensures that the kernel information of each point is preserved; 2) unlike EdgeConv, which only uses point coordinates as its features, we additionally take intensity into consideration by concatenating it after each local feature embedding layer; and 3) in \cite{Wang2019DynamicGC} the authors directly replace the fully connected layers in PointNet \cite{Qi2017PointNetDL} architecture with EdgeConv layers, while we additionally propose a front block to embed local features first. Experimental results show that our proposed method achieves better performance on 3D object detection tasks.

\begin{figure}[t]
\centering
\includegraphics[width=0.88\linewidth]{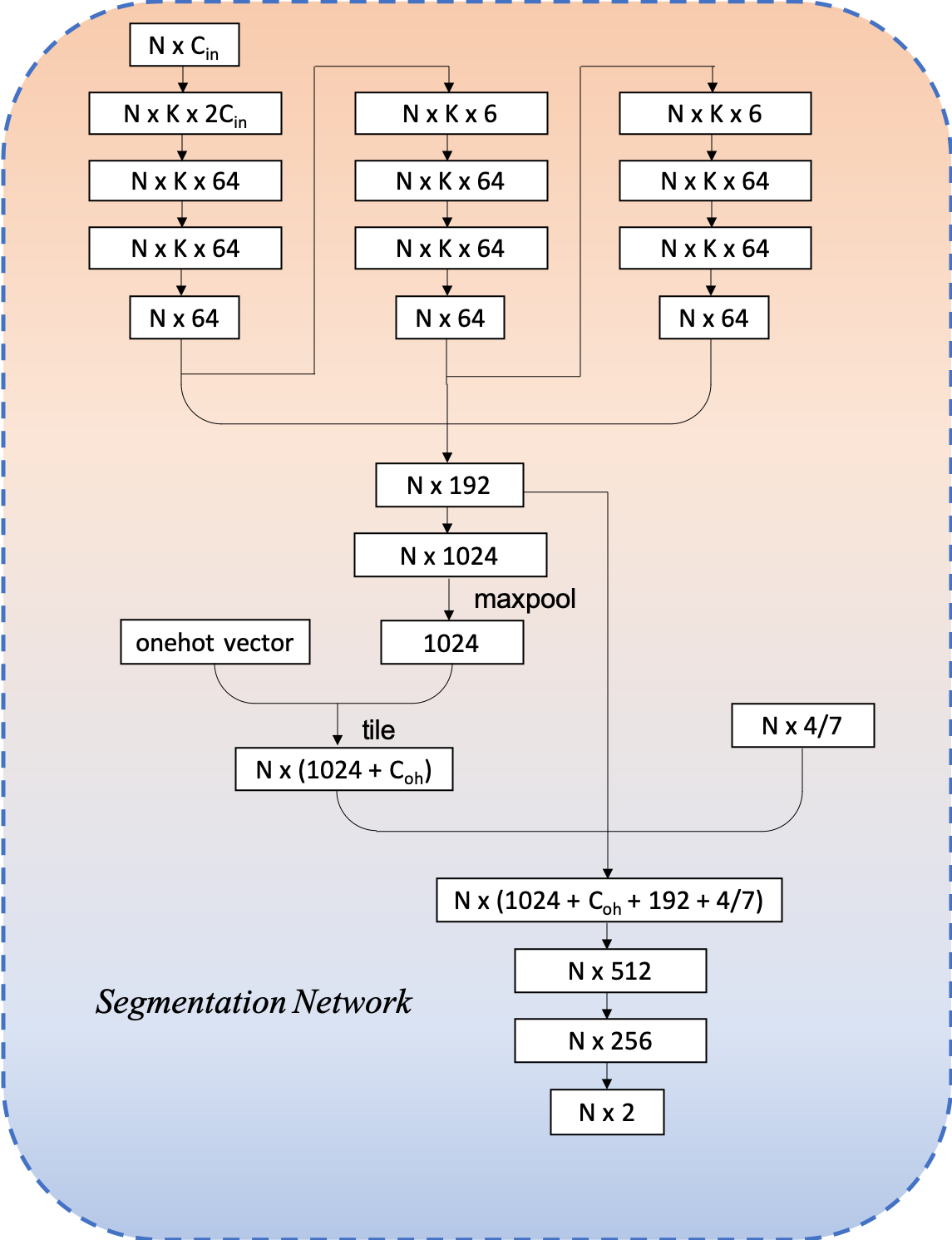}
    \caption{Structure of our segmentation network. Input may be the point clouds directly or the ones processed through the embedding block. Additional features obtained from the spatial transform net are concatenated before the last three layers.}
\label{fig:segNet}
\end{figure}

\renewcommand{\arraystretch}{1.1}
\begin{table*}[t]
\begin{center}
\scalebox{1.3}{
\begin{tabular}{c||ccc||ccc||ccc}
\hline
\multirow{2}{*}{Method} & \multicolumn{3}{c||}{Cars} & \multicolumn{3}{c||}{Pedestrians} & \multicolumn{3}{c}{Cyclists} \\
\cline{2-10} 
& Easy & Moderate & Hard & Easy & Moderate & Hard & Easy & Moderate & Hard \\
\hline
F-PointNet \cite{Qi2018frustum} & 80.62 & 64.70 & 56.07 & 50.88 & 41.55 & 38.04 & 69.36 & 53.50 & 52.88 \\
F-PointNet++ \cite{Qi2018frustum} & 81.20 & 70.39 & 62.19 & 51.21 & 44.89 & 40.23 & 71.96 & 56.77 & 50.39 \\
DGCNN \cite{Wang2019DynamicGC} & 84.99 & 71.58 & 63.95 & 67.71 & 58.25 & 51.16 & 70.51 & 53.03 & 49.91 \\
\hline
Ours (EB)       & 84.97 & 72.61 & 64.64 & 68.35 & 59.19 & 51.79 & 79.15 & 59.25 & 56.24 \\
Ours (EB+FCR)   & \textbf{85.30} & \textbf{72.66} & \textbf{64.72} & \textbf{69.01} & \textbf{60.15} & \textbf{52.89} & \textbf{83.38} & \textbf{62.05} & \textbf{57.81} \\
\hline
\end{tabular}
}
\end{center}
\caption{3D object detection average precision (AP) on KITTI test set. 3D bounding box intersection over union (IoU) threshold is $70\%$ for cars and $50\%$ for pedestrians and cyclists.}
\label{table:detection}
\end{table*}

\begin{table*}[t]
\begin{center}
\scalebox{1.3}{
\begin{tabular}{c||ccc||ccc||ccc}
\hline
\multirow{2}{*}{Method} & \multicolumn{3}{c||}{Cars} & \multicolumn{3}{c||}{Pedestrians} & \multicolumn{3}{c}{Cyclists} \\
\cline{2-10} 
& Easy & Moderate & Hard & Easy & Moderate & Hard & Easy & Moderate & Hard \\
\hline
F-PointNet \cite{Qi2018frustum} & 87.28 & 77.09 & 67.90 & 55.26 & 47.56 & 42.57 & 73.42 & 59.87 & 52.88 \\
F-PointNet++ \cite{Qi2018frustum} & 88.70 & 84.00 & 75.33 & 58.09 & 50.22 & 47.20 & 75.38 & 61.96 & 54.68 \\
DGCNN \cite{Wang2019DynamicGC} & 88.39 & 83.01 & 75.88 & 71.60 & 62.39 & 57.14 & 77.60 & 59.82 & 55.48 \\
\hline
Ours (EB)       & 88.30 & \textbf{84.13} & 76.20 & 71.87 & 65.58 & 58.45 & 84.88 & 65.15 & \textbf{61.14} \\
Ours (EB+FCR)   & \textbf{88.46} & 83.83 & \textbf{76.50} & \textbf{72.15} & \textbf{66.16} & \textbf{58.80} & \textbf{86.14} & \textbf{65.49} & 60.81 \\
\hline
\end{tabular}
}
\end{center}
\caption{3D object localization AP (bird's eye view) on KITTI test set. 3D bounding box IoU threshold is $70\%$ for cars and $50\%$ for pedestrians and cyclists.}
\label{table:birdView}
\end{table*}

\subsection{Using Point-Neighborhood Embedding for the\\ Segmentation Network}
The baseline of our work, F-PointNet, used PointNet as its segmentation network. It has also been used in numerous other papers as their backbone network architecture. In this paper, we also adopt the PointNet structure but with some modifications for our segmentation network.
There are two possible approaches to apply our proposed KNN-based point embedding operation. One is the aforementioned method of having all the points pre-embedded through an embedding network, and then process the embedded features through a normal segmentation network, e.g. PointNet \cite{Qi2017PointNetDL}. The other approach is to directly replace the initial fully connected layers in the segmentation network with point neighborhood embedding operations without adding an additional embedding block. This replacement is only applied in the early stages. The last layers in the segmentation network are high-dimensional regression fully connected layers. Replacing them with embedding operations is not considered here as the computational requirements will increase substantially. See Figure \ref{fig:segNet} for the whole segmentation network architecture after the layer replacement. Combinations of these two approaches are also possible. Although the second approach appears more common and promising at first glance, for the specific frustum-based 3D object detection task, our experiments show that the first method actually contributes more to the performance improvements. See details in Section~\ref{sec:experiments}.

In our network structure, additional features including one-hot vector labels, original point coordinates, and transformed point coordinates are concatenated to different intermediate step features before the final step of regression. Note that we also additionally consider the skip connections that have been widely used after the proposal of ResNet architecture \cite{He2016ResNet}. These skip connections can keep the features at different layers in a combing way for next step computations, which enable the network to get the features at different levels for a better learning.

To conclude, Table \ref{table:method comparison} gives a full comparison between our method and competing methods from the literature regarding input processing operator $\mathbb{D}$ and other sub-blocks.

\vspace{5pt}
\section{Experiments}
\label{sec:experiments}

\subsection{Datasets and Implementation Details}
The evaluation dataset we used in this paper is the KITTI dataset \cite{Geiger2012AreWR}, which is one of the largest computer vision algorithm evaluation dataset in the world. Its 3D object detection dataset consists of 7481 training examples and 7581 testing examples, which are extracted from a number of sequences from autopilot scenarios. Each example contains two RGB color images from left and right stereo cameras as well as the corresponding point cloud frames captured by a Velodyne laser scanner. KITTI contains real-world image data from scenes such as urban, rural, and highways, with up to 15 vehicles and 30 pedestrians per images, as well as varying degrees of occlusion and truncation.

In this paper we follow the framework of F-PointNet with exception of the segmentation part. In the segmentation part, we choose $K=4$ when applying the KNN-based local feature embedding operations, which means 4 nearest neighbor points are considered for each point. In the embedding operations, we choose $\Cout=64$ as the intermediate and output dimensionality of the embedded feature map them. Same as the output dimensionality in the fully connected layer replacement method. Ablation explanation of these choices are given in Section \ref{subsec:ablation studies}. We train our network with a mini-batch size of 32 on one GPU, using the ADAM optimizer. Learning rate starts from 0.001 with a decay of a factor 0.5 every 800,000 iterations (about 10 epochs) during a total of 200 training epochs.

\subsection{Comparisons with Baseline}
We use a identical evaluation method and performance metrics of the proposed method to the evaluation of Frustum-PointNet. Performance on KITTI validation dataset in terms of 3D average precision and top-view average precision is presented in Table \ref{table:detection} and Table \ref{table:birdView}, respectively. For easy comparison, the result tables also include the baseline results from Frustum-PointNet, both the one used PointNet as the backbone structure and the one used PointNet++ (we refer it as F-PointNet++).

From Table \ref{table:detection} and Table \ref{table:birdView} it is clear that our method outperforms the baseline Frustum-PointNet on all detection tasks. Notably, the method of fully connected layers replacement (FCR) as used in \cite{Wang2019DynamicGC} also improves the detection performance, but does not contribute as much as ours. 

\begin{figure*}[t]
\centering
\subfigure[False negative corrected.]{
    \includegraphics[width=0.9\linewidth]{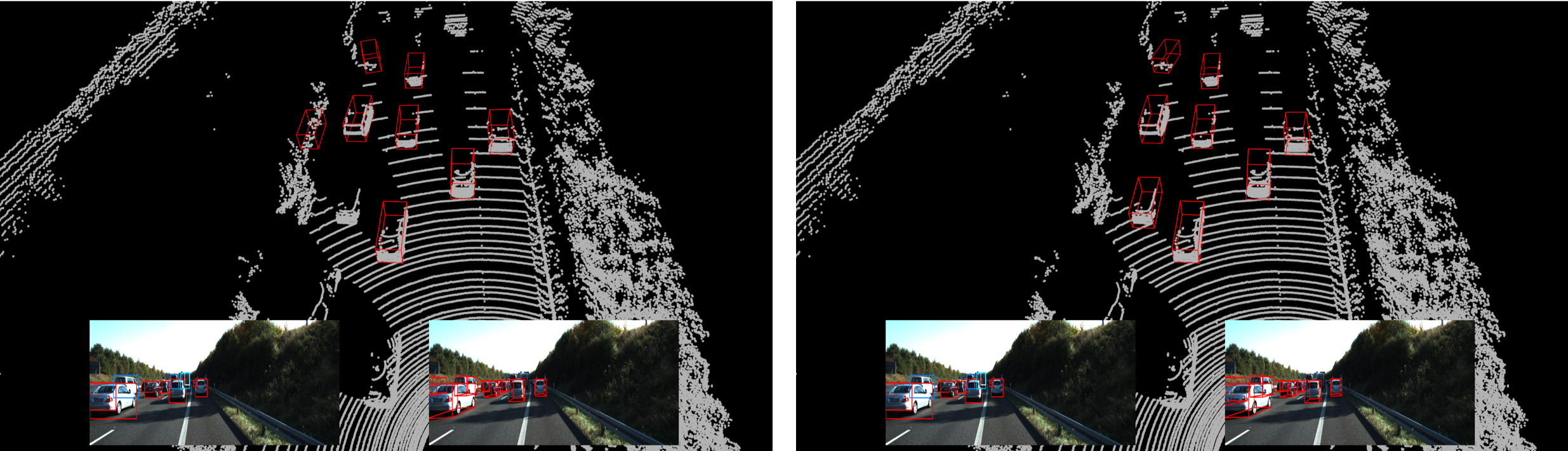}}
    \label{fig:FN}
\subfigure[False positive corrected.]{
    \includegraphics[width=0.9\linewidth]{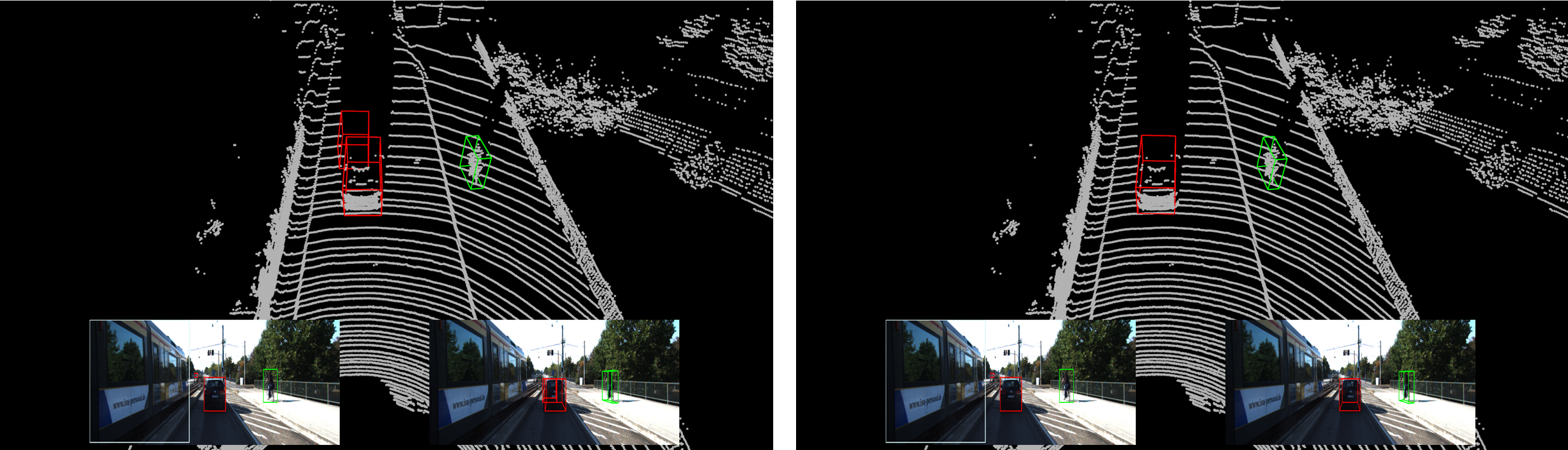}}
    \label{fig:FP}
\subfigure[Bounding box improved, example 1.]{
    \includegraphics[width=0.9\linewidth]{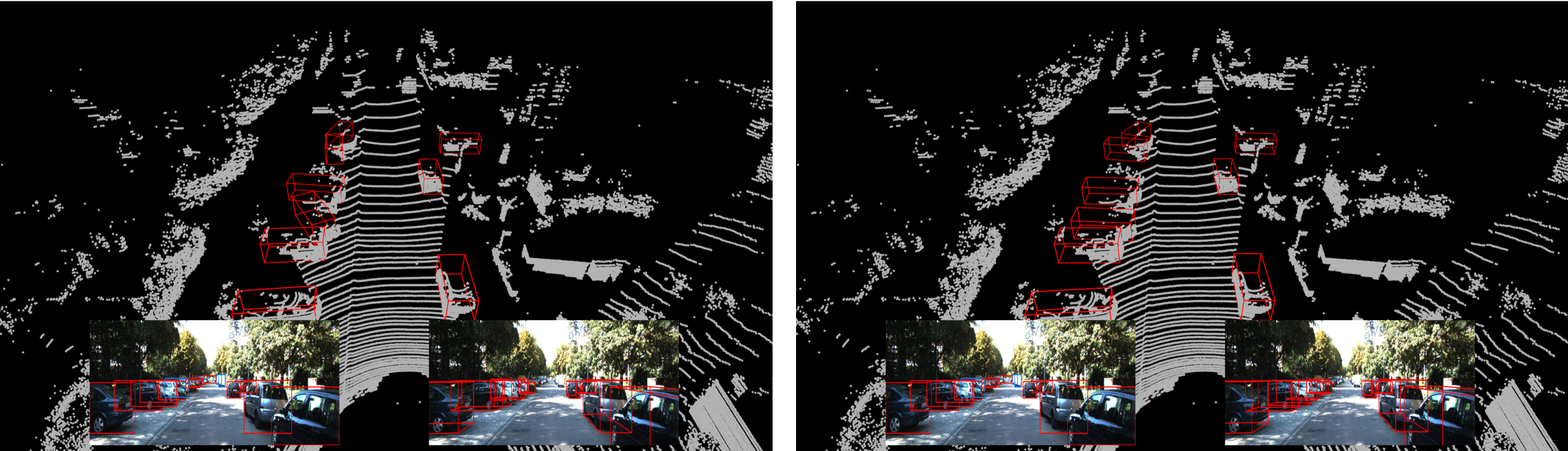}}
    \label{fig:Im1}
\subfigure[Bounding box improved, example 2.]{
    \includegraphics[width=0.9\linewidth]{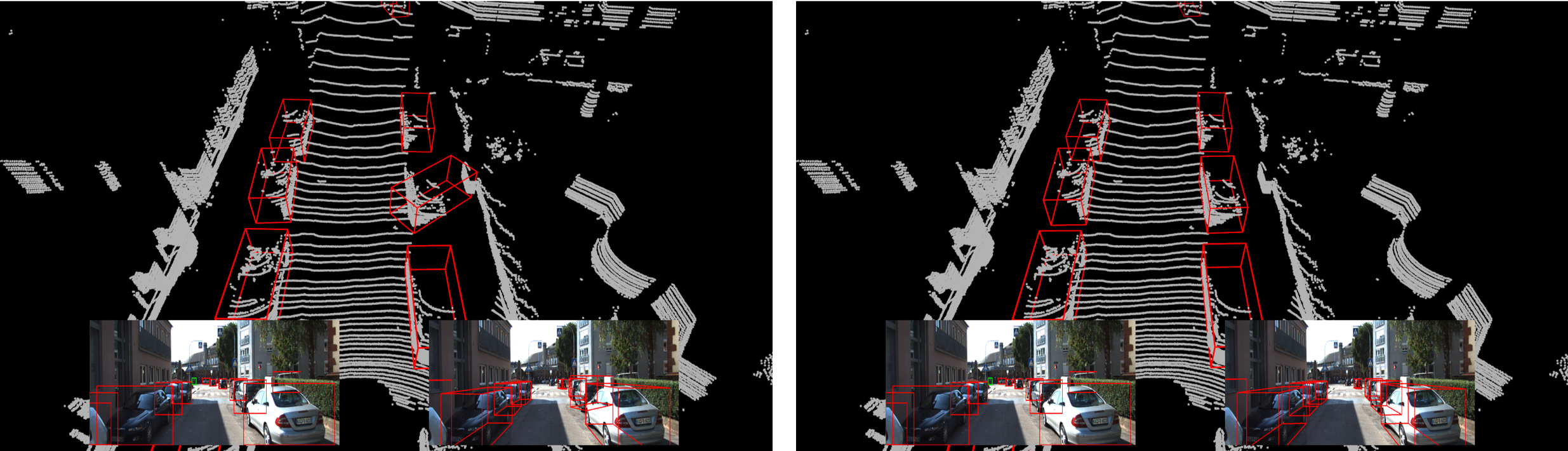}}
    \label{fig:Im2}
\caption{Comparison examples between the results of our method and the results obtained from Frustum-PointNet. Detection results have been corrected or improved under different situations.}
\label{fig.comparison}
\end{figure*}

Some visualized qualitative results have been given in Figure \ref{fig.comparison}. Comparing to the results with F-PointNet, our method is able to correct possible false negative and false positive errors, as well as to improve the quality of 3D bounding boxes by correcting the box orientations.

\begin{table}[t]
\begin{center}
\scalebox{1.3}{
\begin{tabular}{c|c|c||ccc}
\hline
\multicolumn{3}{c||}{Method} & \multicolumn{3}{c}{Cars} \\ 
\hline
\multicolumn{2}{c|}{Embedding} & \multirow{2}{*}{FCR} & \multirow{2}{*}{Easy} & \multirow{2}{*}{Moderate} & \multirow{2}{*}{Hard} \\ \cline{1-2}
ST & LFE &  &  &  & \\  
\hline
- & - & - & 81.20 & 70.39 & 62.19 \\
- & - & \checkmark & 84.99 & 71.58 & 63.95 \\
\checkmark & - & - & 84.64 & 72.24 & 64.29 \\
- & \checkmark & - & 84.53 & 71.14 & 63.50 \\
\checkmark & \checkmark & - & 84.97 & 72.61 & 64.64 \\
\checkmark & \checkmark & \checkmark & \textbf{85.30} & \textbf{72.66} & \textbf{64.72} \\
\hline
\end{tabular}
}
\end{center}
\caption{Sub-block test experiments. ST stands for the spatial transformation sub-block, LFE stands for the local feature embedding sub-blocks, FCR  stands for the fully connected layers replaced.}
\label{table:subBlocks}
\end{table}

\begin{figure}[t]
\begin{center}
    \includegraphics[width=1.0\linewidth, height=3.5cm]{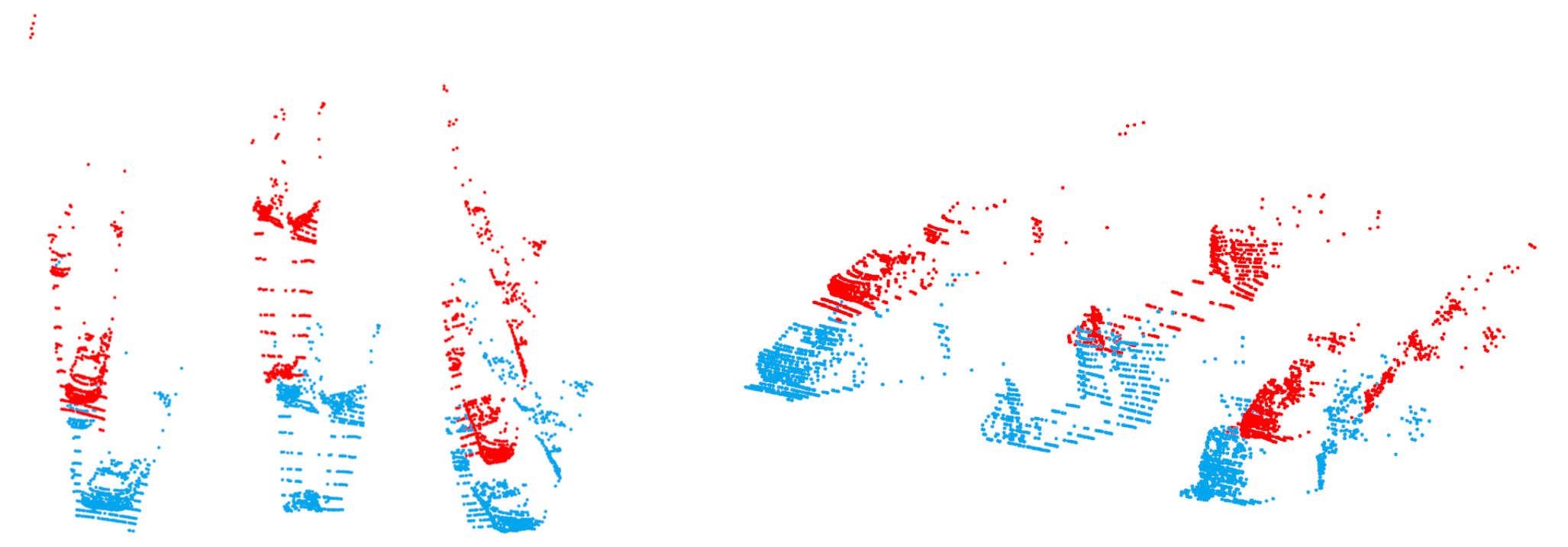}
\end{center}
    \caption{Visualization of example frustums after the spatial transformation sub-block, red and blue point clouds are the frustums processed before/after the ST net, respectively. Left: top view. Right: side view. From the top view we can clearly observe that frustums have been slightly rotated to similar orientations, i.e. left frustums slightly rotate right, while right ones rotate left.}
\label{fig:STNvis}
\end{figure}

\subsection{Ablation studies} 
\label{subsec:ablation studies}
In order to better illustrate how each sub-block contribute to the final performance, multiple experiments have been performed on different combinations of these sub-blocks, as shown in Table \ref{table:subBlocks}. It shows that the spatial transform net contributes the most in our method (note that the local features have already been considered in the ST net). With the full embedding block applied, additionally replacing the fully connected layers with the local embedding layers only marginally improves the performance. 

To give a better illustration of how the ST sub-block works, we visualized the transformed frustums after spatial transformation, examples are shown in Figure \ref{fig:STNvis}. From them we may observe that almost all frustums are slightly rotated to similar orientations. At the same time, it is also noticeable that all  frustums are slightly compressed along the viewing direction, a possible explanation is that this is a weak normalization operation. Following this idea, dividing the frustums into more cubic parts to perform normalization partially may further improve the performance.

\begin{figure}[t]
\begin{center}
    \includegraphics[width=1.0\linewidth]{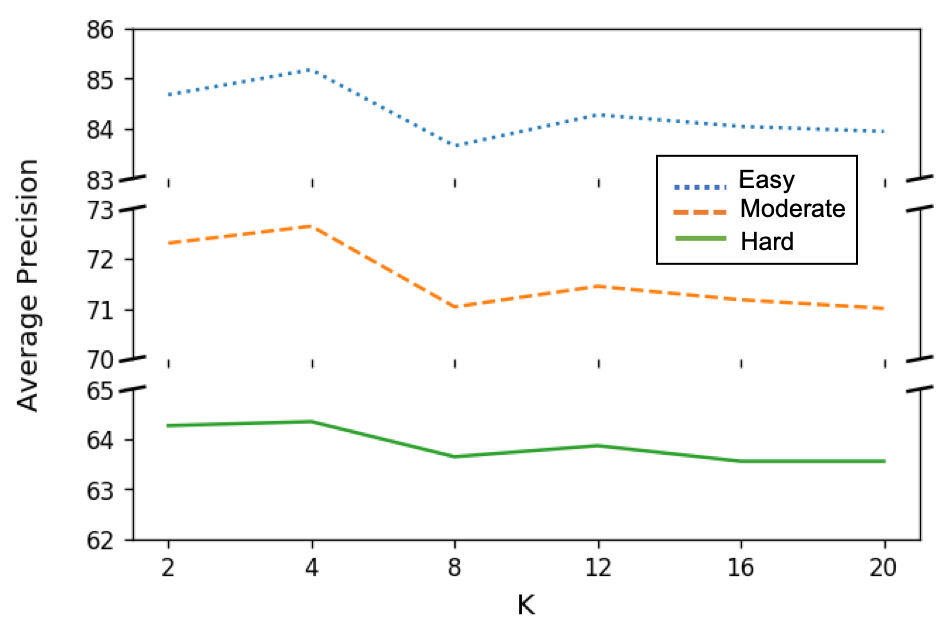}
\end{center}
    \caption{3D object detection AP on KITTI test set for cars with different selection of $K$. A choice of $K=4$ or $5$ is the best setting in our experiments.}
\label{fig:K}
\end{figure}

In this paper, we choose $K=4$ for the KNN searching, while \cite{Wang2019DynamicGC} reports a choice of $K=20$. We tested our method with different settings of $K$, results are shown in Figure \ref{fig:K}. Although all performances are better than the F-PointNet baseline, it is clear that for our tasks increasing $K$  does not further improve performance. However, it is also possible to use different $K$ at different layers of the network. With delicate choices of $K$ at different layers, the performance may also be improved with further experiments.

\vspace{5pt}
\section{Conclusion}
In this paper, based on one of the state-of-the-art 3D object detection methods, Frustum-PointNet, we have proposed an improved version by leveraging local correlations in point clouds. A KNN-based local information gathering operation is applied to embed local features of points. After performing spatial invariance transformation and local feature embedding, the initial feature of point clouds are expanded to a higher dimensionality with local correlations embedded. Experimental results show that with our newly proposed method, better performance on 3D object detection tasks is achieved compared to the original F-PointNet baseline. 
For future works, we are planning to: (i) design other possible pre-processing operations for the better normalization of input frustums, (ii) take the non-Euclideaness of point cloud data into consideration when computing its local feature, and (iii) construct more flexible and interpretable network architectures.

\vspace{1cm}
\bibliographystyle{unsrt}
\bibliography{bibfile}

\end{document}